%% file: main.tex
\documentclass[10pt,twocolumn,letterpaper]{article}

\usepackage{iccv}
\usepackage{times}
\usepackage{epsfig}
\usepackage{graphicx}
\usepackage{amsmath}
\usepackage{amssymb}

\usepackage{color}
\usepackage{booktabs}
\usepackage{multirow}
\usepackage{xcolor,colortbl}
\usepackage{stfloats}
\usepackage{soul}
\usepackage{ulem}

\usepackage{balance}

\usepackage[pagebackref=true,breaklinks=true,letterpaper=true,colorlinks,bookmarks=false]{hyperref}

\iccvfinalcopy 

\def\iccvPaperID{2199} 

\ificcvfinal\fi

\begin{document}

\title{Open Vocabulary Multi-Label Classification with \\Dual-Modal Decoder on Aligned Visual-Textual Features}

\author{Shichao Xu\\
Northwestern University\\
Evanston, USA\\
{\tt\small shichaoxu2023@u.northwestern.edu}
\and
Yikang Li\\
OPPO US Research Center\\
Palo Alto, USA\\
{\tt\small yikang.li1@oppo.com}
\and
Jenhao Hsiao\\
OPPO US Research Center\\
Palo Alto, USA\\
{\tt\small mark@oppo.com}
\and
Chiuman Ho\\
OPPO US Research Center\\
Palo Alto, USA\\
{\tt\small chiuman@oppo.com}
\and
Zhu Qi\\
Northwestern University\\
Evanston, USA\\
{\tt\small qzhu@northwestern.edu}
}

\maketitle
\ificcvfinal\thispagestyle{empty}\fi

\begin{abstract}
In computer vision, multi-label recognition are important tasks with many real-world applications, but classifying previously unseen labels remains a significant challenge.
In this paper, we propose a novel algorithm,  \textbf{A}ligned \textbf{D}ual mo\textbf{D}ality  Cla\textbf{S}sifier (ADDS), which includes a Dual-Modal decoder (DM-decoder) with alignment between visual and textual features, for open-vocabulary multi-label classification tasks. Then we design a simple and yet effective method called Pyramid-Forwarding to enhance the performance for inputs with high resolutions. Moreover, the Selective Language Supervision is applied to further enhance the model performance.
Extensive experiments conducted on several standard benchmarks, NUS-WIDE, ImageNet-1k, ImageNet-21k,
and MS-COCO, demonstrate that our approach significantly outperforms previous methods and provides state-of-the-art performance for open-vocabulary multi-label classification, conventional multi-label classification and an extreme case called single-to-multi label classification where models trained on single-label datasets (ImageNet-1k, ImageNet-21k) are tested on multi-label ones (MS-COCO and NUS-WIDE).

\end{abstract}


\input{sections/introduction}

\input{sections/related_work}
\input{sections/method}
\input{sections/experiment}

\input{sections/conclusion}

{\small
\bibliographystyle{ieee_fullname}
\bibliography{egbib}
\balance
}

\end{document}

%% file: sections/introduction.tex
\section{Introduction}
Image classification tasks are fundamental and critical in computer vision. In recent years, with the advancement of deep neural networks, image classification has made a great impact in many real-world applications, such as medical imaging~\cite{li2014medical}, autonomous driving~\cite{fujiyoshi2019deep}, manufacturing~\cite{rendall2018image}, agriculture~\cite{kamilaris2018deep}, etc. 
The most prevalent and well-studied topic in image classification is the \textit{single-label classification}~\cite{deng2009imagenet, ridnik2021imagenet}, which assumes that each image contains only one item, scene, or concept of interest to be labeled. Then, the topic of conventional \textit{multi-label classification}~\cite{yang2016exploit, wang2017multi, gao2021learning} is brought out to address the case where multiple objects, scenes, or concepts in the same image are of interest and need to be labeled.

However, the conventional multi-label classification still cannot meet the need of some real-world applications where unseen labels may occur during testing. Previously, the task with unseen labels is often referred to as multi-label zero-shot learning. 
The related methods are often created based on label correlations~\cite{lee2018multi, lu2020multi}, which try to identify potential relations among the labels within the image to facilitate classification, usually, through building a label graph. Moreover, some methods~\cite{ben2021semantic, liu2021query2label}, including the previous state-of-the-art (SOTA) method ML-Decoder~\cite{ridnik2021ml}, create label embeddings solely from the words or assign learnable embeddings for each label instead of indirectly creating and learning from the graph relation, allowing for more sophisticated outcomes.
One of the mostly used label embeddings is Word2Vec~\cite{mikolov2013efficient}, which is created from pretraining tasks with external textual datasets (target label classes can be seen during pre-training). Then by extracting local discriminative features according to different label embeddings, the model outputs the per-class probability.
However, these models only focus on the connection between the words, and the generalization of the learned mapping (from image to seen label) to the target mapping (from image to unseen label) is still challenging. The leveraging of the word embedding also blocks the model from handling phase/sentence labels, which also exist in practice and are very challenging to address.

In~\cite{zareian2021open}, the \textit{Open-Vocabulary} setting is introduced, which is a generalization of zero-shot and weakly supervised settings and is more suitable for dealing with unseen classes. While the target classes are not known during training, it can be any subset of the entire language vocabulary in the pretraining tasks (e.g., contrastive learning on image-caption dataset). It is proved to be quite effective in some computer vision tasks, such as object detection~\cite{du2022learning} and object segmentation~\cite{ghiasi2021open}. In this setting, instead of using costly annotations for classification datasets, the Vision-Language Pretraining (VLP) model trained on image-caption datasets can be utilized to help build the connection between the visual and textual embeddings and provide more flexibility in algorithm design.

However, current VLP models are not silver bullets and present new challenges. Practical VLP models are usually trained with fixed low-resolution images for reducing the computational cost of large-scale data sources. The input resolution for the model adapted from those VLP models will be restricted. Besides, identifying whether a label exists in the image from their embeddings is also challenging, as the measurement based on simply comparing the cosine similarity can lead to an uncertain threshold (which is typically different for different images and objects.) 
Besides, given the difficulty and cost to collect and annotate multi-label datasets, we also consider a somewhat extreme but practically useful setting where models trained on single-label datasets (e.g., ImageNet-1k) are tested on multi-label datasets with unseen categories (e.g., NUS-WIDE). We call this \textit{single-to-multi label classification}. 


In this work, we develop a novel approach for open-vocabulary multi-label classification that significantly outperforms previous methods and provides the new state-of-the-art (SOTA) performance. 
In three classification tasks -- open-vocabulary multi-label, single-to-multi label, and conventional multi-label  -- our approach provides significantly higher mean Average Precision (mAP) than prevoius methods, including SSGRL~\cite{chen2019learning}, MS-CMA~\cite{you2020cross}, ASL~\cite{ben2020asymmetric}, Q2L~\cite{liu2021query2label},
  LESA~\cite{huynh2020shared}, BiAM~\cite{narayan2021discriminative}, GMLZSL~\cite{gupta2021generative}, SDL~\cite{ben2021semantic}, ML-Decoder~\cite{ridnik2021ml}.
More specifically, to overcome the challenges in previous methods, we propose an open-vocabulary multi-label classification framework \textbf{ADDS} (\textbf{A}ligned \textbf{D}ual mo\textbf{D}ality  Cla\textbf{S}sifier) based on the aligned visual and textual embeddings. The framework includes a novel \textbf{DM-decoder} (Dual-Modal decoder) design, which leverages the dual modality to enhance transformer decoder layers by progressively fusing visual embeddings with textual information and developing richer semantic understanding. It also includes a  \textbf{Pyramid-Forwarding} method to adapt the model pre-trained on lower image resolutions to higher resolution images without re-training.   
To summarize, our work makes the following technical contributions:
\begin{itemize}
    \item We have developed ADDS, an open-vocabulary multi-label classification framework that builds on aligned visual and textual features. The framework includes DM-Decoder, a novel transformer decoder for facilitating the fusion of the semantics from dual-modal information source, and Pyramid-Forwarding, a new adaptation method that addresses images with higher resolutions than training images and is also able to largely reduce the computational cost of vision transformer.
    \item We have conductecd extensive experiments across various multi-label classification tasks. Our ADDS framework significantly outperforms the previous SOTA methods in all scenarios, including open-vocabulary multi-label classification ($11.57$ points improvement on NUS-WIDE), single-to-multi label classification ($24.71$ points and $16.49$ points improvements from ImageNet-1k to MS-COCO and NUS-WIDE, respectively), and conventional multi-label classification (e.g., $2.14$ points improvement on MS-COCO).
\end{itemize}

%% file: sections/related_work.tex
\section{Related Works}
\subsection{Conventional Multi-label Classification}
Conventional multi-label classification, which aims to classify multiple objects, scenes, or concepts in a given image, is generally a more challenging task than the typical single-label classification. It has been studied in the literature by various approaches.
The first group of methods is based on the region of interest. And in previous works such as~\cite{yang2016exploit, wang2017multi, gao2021learning, you2020cross, gao2021learning}, multi-label classification is solved by locating each object in the image or capturing the attention map and then performing single-label classification on it. 
However, these methods often suffer from issues like coarse discovered regions, heavy computation costs, some concepts or scenes being hard to localize, and some regions containing duplicate concepts.
Another group of methods is based on label correlations. They try to identify the potential relations among the labels within the training images to facilitate classification~\cite{chen2019learning, shi2020multi, ye2020attention}. For instance,
the method in~\cite{chen2019learning} splits the feature representation into category semantics-specific representations and applies a graph neural network to explore the interactions among them.
Some previous multi-label zero-shot learning methods also share a similar idea with conventional multi-label classification, which will be discussed in the next section.

\subsection{Multi-Label Zero-Shot Classification}
Some of the methods for conventional multi-label classification claim that they can also address zero-shot classification, such as~\cite{yang2016exploit, ridnik2021ml}. There are also other previous works~\cite{norouzi2013zero, akata2015label, zhang2016fast, kim2018bilinear, huynh2020shared, narayan2021discriminative}.
Generally speaking, many papers in recent years try to capture the unseen labels by exploring the connections between the labels.
For instance, Akata et al.~\cite{akata2015label} consider each class as an embedding in
the space of attribute vectors, and then introduce a function measuring the compatibility between an image and a label embedding to determine the correct classes.
The work in~\cite{zhang2016fast} studies the image-word relevance by estimating the principal direction for an image, which is based on the assumption that the word vectors of relevant labels for a given image can rank ahead of the irrelevant labels along a principal direction in the word vector space.
The most recent paper in multi-label zero-shot learning is~\cite{ridnik2021ml}, which employs Word2Vec to generate the label text embedding and solely relies on the relation between the text features for learning. 
However, due to a lack of supervision on the visual information during the textual embedding learning, the learned mapping between the image and text is hard to be generalized to unseen data space.

\subsection{Vision-Language Pre-training (VLP)}

The vision language pre-training learns the semantic correspondence between image and language by pretraining on large-scale data with different tasks. In the literature, some works such as VisualBERT~\cite{li2019visualbert}, Unicoder-VL~\cite{li2020unicoder}, and ViLT~\cite{kim2021vilt} extract image tokens from the interest regions, combine them with language tokens together as the inputs, and fuse the information by the transformer in the early stage. Other works such as Contrastive LanguageImage Pre-training (CLIP)~\cite{radford2021learning}, Self-supervision meets Language-Image Pre-training (SLIP)~\cite{mu2021slip}, Bootstrapping Language-Image Pre-training (BLIP)~\cite{li2022blip}, and Triple Contrastive Learning (TCL)~\cite{yang2022vision} first extract the deep features of the image and the text individually, and then conduct modality interaction after the feature extraction. In this paper, we mainly maintain the alignment of the visual and textual embeddings through CLIP~\cite{radford2021learning}, which is built on the cosine similarity between the image and text embedding pairs and trained with a large and noisy dataset. Moreover, later in Section~\ref{sec:additional_experiments}, we also introduce experiments on using other VLP models such as BLIP~\cite{li2022blip} and SLIP~\cite{mu2021slip}.

\subsection{Open-Vocabulary Learning}
VLP models enable a strong connection between images and corresponding textual information by learning from large-scale training corpora.
Incorporating VLP models into model design has made arbitrary text label prediction possible, and numerous related applications have benefited from it.
In recent years, open-vocabulary object detection~\cite{zareian2021open, gao2022open, gu2021open, bravo2022localized, kuo2022f} and open-vocabulary semantic segmentation~\cite{ghiasi2021open, ma2022open, ghiasi2022scaling} have become increasingly popular. 
Zareian et al.~\cite{zareian2021open} firstly propose Open-Vocabulary object Detection (OVD) and connect it with image-text pretraining. 
Gao et al.~\cite{gao2022open} leverage the localization ability of the VLP model to generate pseudo bounding-box labels for training the open-vocabulary object detector.
Besides, Gu et al.~\cite{gu2021open} further distill the knowledge from a VLP model into a two-stage open-vocabulary object detector.
When VLP models are leveraged for visual-semantic alignments on pixel-level information, Ma et al.~\cite{ghiasi2021open} are able to make the zero-shot transfer to segment novel categories.
In this paper, we utilize the CLIP model to keep the visual-semantic alignments to achieve open vocabulary multi-label classification.

%% file: sections/method.tex
\section{Methodology}

\subsection{Overview}

\begin{figure*}[htb]
\centering
\vspace{-0.2cm}
\includegraphics[width=17cm]{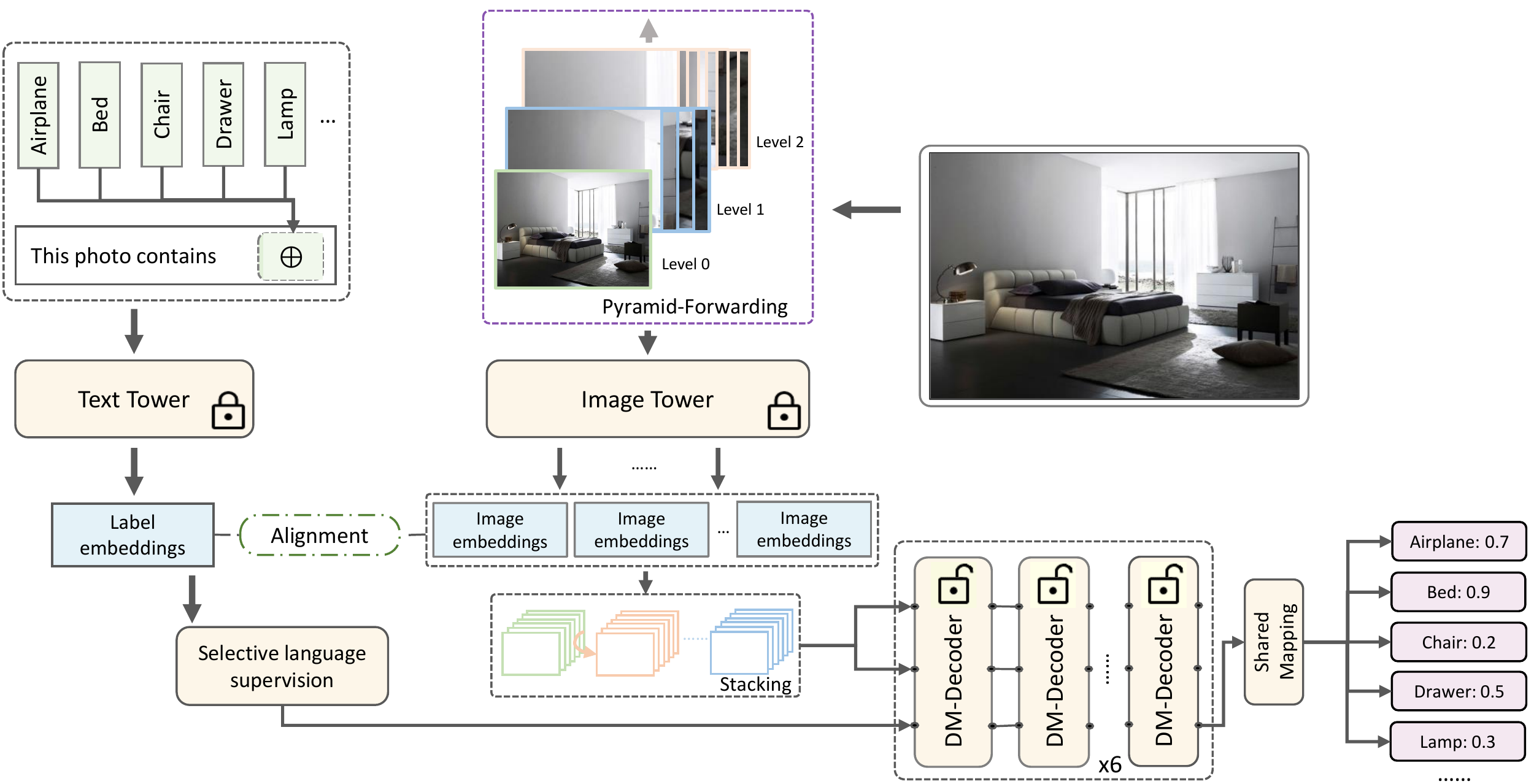}
\caption{Overview of our ADDS framework for multi-label classification. 
Text labels with prompts are fed into a text tower to get the textual embedding. Images are first processed by a Pyramid-Forwarding module and then fed into an image tower to get the visual embeddings, which are aligned with the textual embedding and stacked on the token size dimension.  
Then the textual embedding (after a selective language supervision module) and the stacked visual embeddings are fused by six layers of DM-decoders with the initial query from textual embedding and the initial key/value from visual embeddings. 
After a shared mapping among all labels, the network outputs the probability for each label class.}
\vspace{-0.2cm}
\label{fig:overview}
\end{figure*}

In this section, we present the details of our ADDS method for multi-label classification. 
As shown in Figure~\ref{fig:overview}, our method receives both the image $x_{img} \in \mathbb{R}^{H \times W \times 3}$ and the class names $X_{lbl} \subset \{$natural language words$\}$ as the inputs, where $X_{lbl}$ contain words of potential labels for identifying the objects (tree, apple, computer, \dots), scenes (sea, sky, underground, \dots) or concepts (small, red, \dots). 
Then the classes names $X_{lbl}$ are combined with prompts such as ``This photo contains $@$'' and ``This is a $@$ photo'', where $@ \in X_{lbl}$, and fed through a text tower to get the textual (class) embedding. The image input is fed through the Pyramid-Forwarding module, whose output images are then fed through an image tower to get the visual (image) embedding. The visual embedding is aligned with textual embedding, and then stacked and forwarded to the DM-decoder, whose outputs are mapped to per-class probabilities via a shared fully-connected layer. 
That is, given the input $\{x_{img}, X_{lbl}\}$, our model outputs $p_{pred} = [p_1, p_2, \dots, p_k]$, where $p_i\in [0,1]$, $k = |X_{lbl}|$. 


Compared with previous works~\cite{liu2021query2label, ridnik2021ml} where the label embedding is learned from limited observations, or based on Word2Vec~\cite{ridnik2021ml} or even randomly initialized matrix~\cite{ridnik2021ml}, a major difference of our approach is that our model is constructed based on the alignment between visual and textual embeddings.
This is not only helpful for conventional multi-label classification, but also critical for boosting the performance of open-vocabulary multi-label classification. 
Specifically, in our setting, the training data contains the images $\{x_{(img,seen)}\}$ and the labels $X_{(lbl, seen)}$. The objective is to learn a classifier $g$ to make predictions on an unseen image $x_{(img,unseen)}$ with unseen categories $X_{(lbl, unseen)}$, i.e.,  
$g(x_{(img,unseen)}, x_{(lbl, unseen)}) \in \{0,1\}$,
$g() = 1$ if $x_{(img,unseen)}$ contains object/scene/concept $x_{(lbl, unseen)}$, and $g() = 0$ otherwise. 
Inspired by VLP, we build the visual-semantic alignment with the help of the pre-trained model from CLIP. Specifically, we employ the  vision transformer (ViT)~\cite{dosovitskiy2020image} network architecture as the image encoder $f_{img}$ and the multi-layer transformer as the text encoder $f_{lbl}$, with the parameters of both encoders from CLIP. They are all frozen during training to maintain the alignment (may lead to worse results if unfreeze). Next, we will introduce the major components in our ADDS method in details.

\subsection{Dual-Modal Decoder}
\begin{figure}[htb]
\centering
\vspace{-0.2cm}
\includegraphics[width=0.47\textwidth]{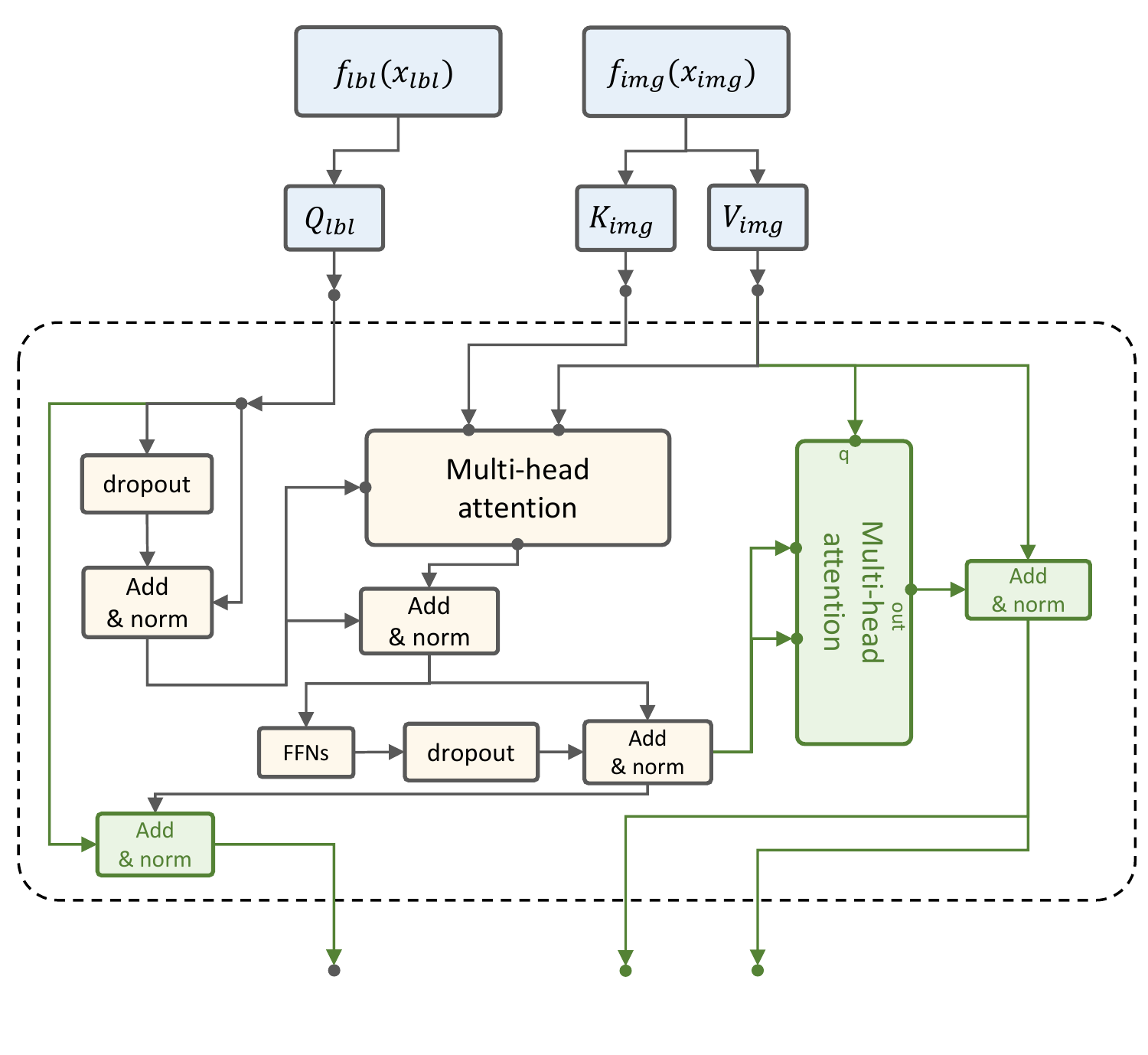}
\vspace{-0.4cm}
\caption{Overview of our Dual-Modal Decoder design.}
\vspace{-0.5cm}
\label{fig:decoder}
\end{figure}

The typical practice of aligned visual and textual embeddings in multi-label classification is to measure their cosine similarity.
Given the visual embedding $E_{img}$ from an image and a textual embedding $E_{lbl}$ from a specific label, whether this image contains the label is determined by the calculated similarity score $s = \cos (\frac{E_{img}}{|E_{img}|}, \frac{E_{lbl}}{|E_{lbl}|})$ and a threshold $\eta$. Usually, the image is deemed to contain the label if $s > \eta$. 

In practice, $\eta$ can be different when considering different images and labels, which presents a significant challenge. Inspired by ML-Decoder~\cite{ridnik2021ml}, this challenge can be mitigated by turning into a binary classification task with adding decoder layers. After the visual and textual feature extraction steps, a single layer cross-attention (we omit the description of the fully-connected layer, dropout, and layer normalization for space reason) is the choice for querying the textual embedding from the visual embedding to determine the per-class probability. 
However, we observe some issues when we stack decoder layers similarly as in~\cite{ridnik2021ml}: 
\begin{itemize}
\item The model performance will often decrease after stacking more than 3 decoder layers. 
\item The key and value inputs are always the same from the visual embedding. As the output of the cross-attention layer is a weighted sum of its value input, the outputs of decoder layers in different levels are actually in the same (or close) semantic level and all from the same visual embedding.
\end{itemize}

To address these issues, we redesign the decoder module, with two major differences from the previous method in~\cite{ridnik2021ml}, 
as shown in Figure~\ref{fig:decoder} and Equation~\eqref{DM-Decoder-eq}. Specifically, we add an additional multi-head cross-attention layer $\textrm{MultiHdAttn}_2$ and use the visual embedding $V_{img}$ to query the output $Q_{mid}^5$ from the previous cross-attention layer $\textrm{MultiHdAttn}_1$, instead of using the textual embedding as the query ($\textrm{MultiHdAttn}_1$ utilizes the textual embedding to query the visual features). The output $Q_{mid}^5$ contains the weighted sum of image tokens' embedding guided by the textual information, so we can redistribute them back to each image tokens' embedding through $\textrm{MultiHdAttn}_2$, to further enhance the visual embedding according to the correlation of $Q_{mid}^5$ with the key inputs. Then after adding and normalization, the key and value input for the next decoder layer are refined by the textual information.
Moreover, apart from the original skipping structures, we add an additional skipping connection from the query input to the query output, which is transformed by addition and normalization. 

Formally, we denote the input query, key, value as $Q_{lbl}, K_{img}, V_{img}$, and the block's output query, key and value for the next block as $Q_{lbl}^{'}, K_{img}^{'}, V_{img}^{'}$. We output $Q_{lbl}^{'}$ only if it is the last layer. We denote $\textrm{DP}$ as the dropout layer, and $\textrm{FFN}_1, \textrm{FFN}_2$ as the fully-connected layer. Then, each new decoder block can be formulated as:
\begin{equation}
\label{DM-Decoder-eq}
\begin{array}{ll}
    Q_{mid}^1 & = \textrm{LayerNorm}(Q_{lbl} + \textrm{DP}(Q_{lbl})),\\
    Q_{mid}^2 & = {\textrm{MultiHdAttn}}_1(Q_{mid}^1, K_{img}, V_{img}),\\
    Q_{mid}^3 & = \textrm{LayerNorm}(Q_{mid}^2 + Q_{mid}^1),\\
    Q_{mid}^4 & = \textrm{DP}(\textrm{FFN}_1 (\textrm{ReLU}(\textrm{FFN}_2(Q_{mid}^3)))),\\
    Q_{mid}^5 & = \textrm{LayerNorm}(Q_{mid}^4 + Q_{mid}^3),\\
    Q_{lbl}^{'} & = \textrm{LayerNorm}(Q_{mid}^5 + Q_{lbl}),\\
    V_{img}^1 & = {\textrm{MultiHdAttn}}_2(V_{img}, Q_{mid}^5, Q_{mid}^5),\\
    V_{img}^{'} & = \textrm{LayerNorm}(V_{img}^1 + V_{img}),\\
    K_{img}^{'} & = V_{img}^{'}.\\
\end{array}
\end{equation}

\subsection{Pyramid-Forwarding}

\begin{figure}[htb]
\centering
\vspace{-0.4cm}
\includegraphics[width=0.48\textwidth]{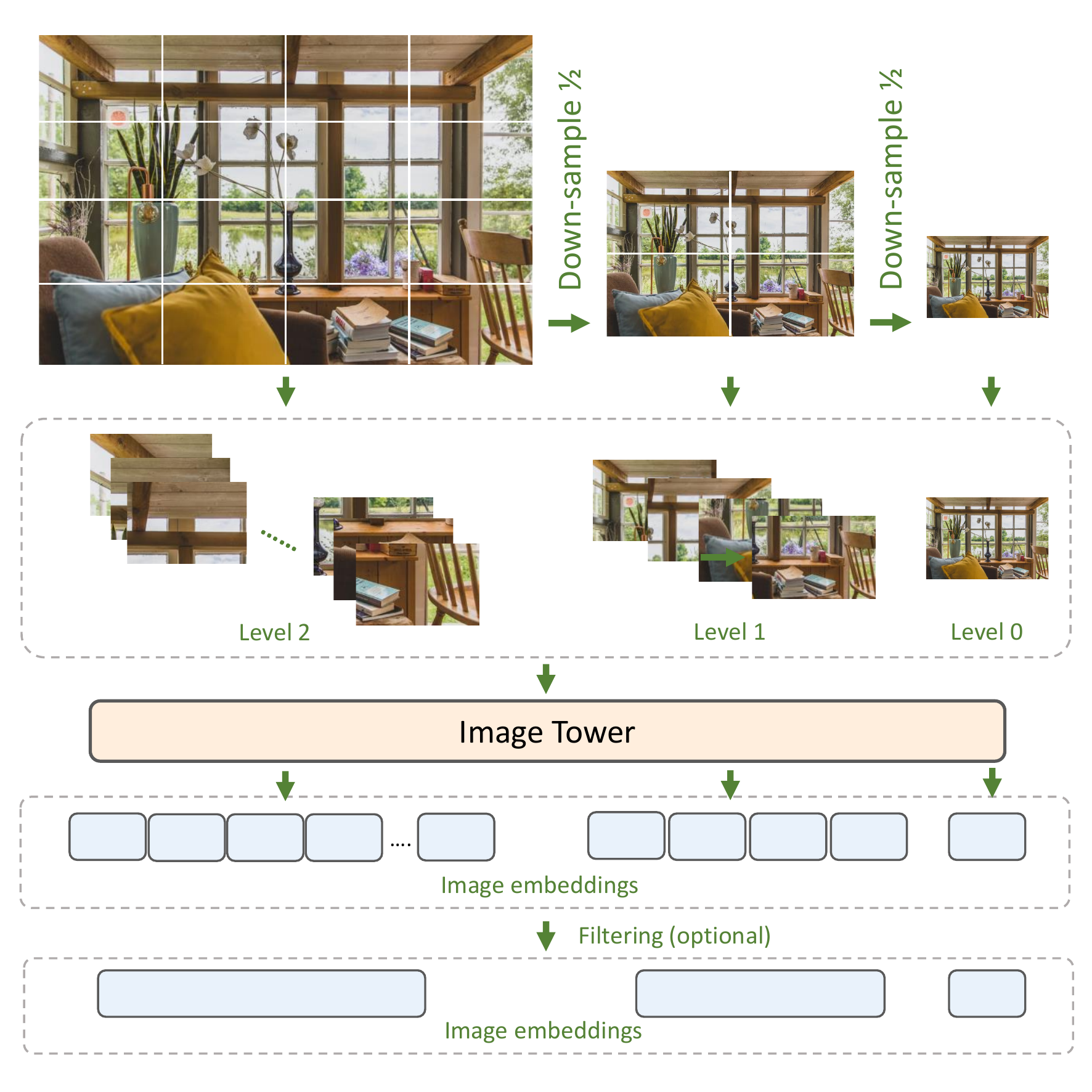}
\vspace{-0.4cm}
\caption{Overview of the Pyramid-Forwarding method.}
\label{fig:infer-5}
\end{figure}

A major challenge for using pre-trained models learned from large-scale datasets (e.g., the VLP models) is that they are often trained on low-resolution images (e.g., 224x224 or 336x336) due to the limitation of computational resources (training time, memory usage, etc.). Thus the extracted features are often not compatible with higher-resolution images.
We could consider downsampling the higher-resolution input images to lower resolution, but that will lose significant local features. Conducting fine-tuning on higher-resolution images might slightly reduce the incompatibility with extracted features, but we still face several challenging questions: 1) Does the model support training on arbitrary scale images? 2) How much fine-tuning data are available to maintain the model's generalization ability? 3) How long or how well the model should be trained to avoid overfitting on the fine-tuning dataset?

To address these challenges, we propose Pyramid-Forwarding, a resolution-compatible method that enables deploying a pre-trained model trained from low-resolution images on high-resolution images without retraining. 
This method applies to many pre-trained image decoders, and below we use the ViT~\cite{dosovitskiy2020image} pre-trained on $S_{Img} \times S_{Img}$ resolution images as an example. It will be applied on $S_{Img}*d \times S_{Img}*d$ target images ($d$ is a positive integer), 
and our method helps reduce the computation cost from $O(d^4 \cdot c)$ to $O(d^2 \cdot c)$, where $c$ is a constant for all resolutions.

Specifically, given a pre-trained model with $S_{Img} \times S_{Img}$ resolution and an image of size $S_{Img}*d \times S_{Img}*d$, Pyramid-Forwarding constructs $\log(d)+1$ levels. First we assume that $\log(d)$ is an integer, and we will discuss the non-integer case later. In level $i \in [0 \dots \log(d)]$, the image is resized to $S_{Img}*2^i \times S_{Img}*2^i$ and split into $(i+1) \times (i+1)$ patches, as visualized in Figure~\ref{fig:infer-5}. Then these patches are fed into the image encoder (usually wrapped into the batch dimension for parallel processing on GPUs) to obtain the feature tokens. In ViT, these tokens are stacked on the token dimension 
and then feed into the decoder, and the computation cost on the $S_{Img}*d \times S_{Img}*d$ image is actually $O((d^2)^2) = O(d^4)$ times more than the $S_{Img} \times S_{Img}$ image because of the self-attention layer. However, with Pyramid-Forwarding, this computation cost is reduced to $1 + 2^2 + 4^2 + \dots + {(2^{\log(d)})}^2 = O(d^2)$. If $\log(d)$ is not an integer, the size of the top-level image will be changed to $S_{Img}^o$, and we allow the divided patches to be overlapped with their neighborhoods, while the total number of patches in this level $i$ is still ${(i+1)}^2$.  
In addition, the computation cost of Pyramid-Forwarding can be further reduced by disposing the image patches from the non-bottom levels $i \in [1 \dots log(d)]$, which provides a trade-off between the accuracy and the computational cost. A special case that works well is when each non-bottom level only contains the token embedding corresponding to the [CLS] token in ViT. It significantly reduces memory usage while moderately degrades the performance, when compared with the complete design. 

\subsection{Selective Language Supervision}
As the label classes can be object/scene/concept names that are selected from the natural language, the total number of the labels can be extremely large (e.g., 1k in ImageNet-1k, 21k in ImageNet-21k, or even more). This presents a challenge in the training stage -- if we feed all the labels into the network, it is a heavy burden on the inference speed and the memory usage. Thus we propose a \textit{selective language supervision} method that utilizes the positive labels and part of the negative labels selected during the network training, and consider the data distribution of the positive and negative samples for reducing data imbalance.

Specifically, given multi-label $L = \{l_1, l_2, \dots, l_k\}$ from the training batch $B$ with $k$ classes in total, $S_{pos}=\{i | l_i = 1, l_i \in L, L \in B\}$, and $S_{neg} = \{1, 2, \dots, k\} - S_{pos}$. 
Then the selected label set for batch $B$ training is 
\begin{equation}
    \vspace{-0.1cm}
    S^{'} = S_{pos} \cup S_{slt},
\end{equation}
where elements in $S_{slt}$ is randomly selected from $S_{neg}$. $|S_{slt}| = \min (\alpha*|S_{pos}|, k - |S_{pos}|)$, where $\alpha$ is a hyper-parameter balancing the number of positive and negative samples (we choose $\alpha = 3$ in our experiments if not mentioned specifically). Note that the selective language supervision is only utilized in experiments with large label set (e.g., the ImageNet-21k dataset).

%% file: sections/experiment.tex
\section{Experimental Results}
\label{sec:experiments}

\subsection{Implementation Settings}
Our code is implemented in PyTorch 1.10.2. All experiments are conducted on a server cluster running Ubuntu 18.04.6 and equipped with NVIDIA V100 GPUs. We conduct the experiments on NUS-WIDE~\cite{chua2009nus}, 
MS-COCO~\cite{lin2014microsoft}, ImageNet-1k~\cite{deng2009imagenet}, and ImageNet-21k~\cite{ridnik2021imagenet} datasets. All models are optimized by the Adam optimizer~\cite{kingma2014adam}. The learning rate is set as $3 \times 10^{-4}$ for all models trained on $224 \times 224$ and $336 \times 336$ images, and as $1 \times 10^{-4}$ for all models trained on images larger than $336 \times 336$. We test the input resolution of the images on $224 \times 224, 336 \times 336, 448 \times 448, 640 \times 640, 1344 \times 1344$, and they are specified in each experiment. We adopt the ViT~\cite{dosovitskiy2020image} as the image encoder and use CLIP~\cite{radford2021learning} pre-trained weights. The output of ViT is $[bs, n_t,l]$, where $bs=56$ is the batch size, $n_t$ is the number of output tokens from the vision transformer, and $l$ is the embedding length determined by the type of the vision transformer. 
All image and text towers in our method are fixed during the training. The model is trained for $40$ epochs, and the weight decay is $10^{-4}$.
Our model applies the loss function ASL from~\cite{ben2020asymmetric}, which is also used in the most recent multi-label classification works~\cite{liu2021query2label, ridnik2021ml}.

\subsection{Open-Vocabulary Multi-label Classification}
In open-vocabulary multi-label classification, we first show our results on the NUS-WIDE dataset in Table~\ref{tab:nus}. The marking `uf' means unfreezing the backbone network and `f' means freezing the backbone network, and both markings are only used when ML-Decoder is switched to the same backbone as our model. To begin with, we use image-encoder ViT-B-32 and its corresponding text encoder with CLIP pre-trained weights. Using $224 \times 224$ input images, our ADDS method already achieves a $36.56\%$ mAP, $5.46$ points higher than the previous SOTA method ML-Decoder with TresNet-L as the backbone network architecture and trained on $448 \times 448$ input images. Then we use image-encoder ViT-L-336 and its corresponding text encoder with CLIP pre-trained weights on $336 \times 336$ images. The result further improves to a $39.01\%$ mAP, $7.9$ points higher than ML-Decoder (which  uses higher resolution images than both of our first two settings). Finally, our result can reach $42.67\%$ when trained on $448 \times 448$ images. We also show that when we use the same backbone network architecture ViT-L-336 and the same pretraining weights (with freezing or unfreezing the backbone) for ML-Decoder, our ADDS method still shows significant advantage (i.e., $39.01\%$ mAP over $33.7\%$ mAP for $336 \times 336$ images; note that ML-Decoder cannot be trained with $448 \times 448$ images for ViT-L-336 as it does not support input resolutions different from the original model resolution). Overall, the results in Table~\ref{tab:nus} clearly demonstrate that \textbf{our approach provides significant improvement over previous methods on open-vocabulary multi-label classification}.

\begin{table}[htb]
\centering
\small
\setlength{\tabcolsep}{1.4mm}{
\begin{tabular}{p{3.1cm}|c|c|c|c}

\hline
Method      & Type              & mAP(\%)                   & F1 (k=3)              & F1 (k=5)              \\ \hline 
CONSE          & zsl           & 9.4                   & 21.6                  & 20.2                  \\ 

LabelEM & zsl & 7.1                   & 19.2                  & 19.5                  \\

Fast0Tag        & zsl           & 15.1                  & 27.8                  & 26.4 \\

One Attention per Label & zsl   & 10.4                  & 25.8                  & 23.6          \\

One Attention per Cluster (M=10) & zsl  & 12.9 & 24.6 & 22.9 \\

LESA     & zsl           & 19.4                  & 31.6                  & 28.7                  \\ 

BiAM       & zsl                & 26.3                  & 33.1                  & 30.7                  \\ 

Generative ML-ZSL   & zsl          & 25.7 & 32.8 & 29.3 \\ 

SDL         & zsl               & 25.9                  & 30.5                  & 27.8                  \\ 

ML-Decoder (TresNet-L, 448x448)    & zsl             & 31.1                  & 34.1                  & 30.8                  \\
ML-Decoder (ViT-L-336, uf, 336x336)   & zsl             & 16.6                  & 16.2                  & 17.8                  \\
ML-Decoder (ViT-L-336, f, 336x336)    & zsl            & 33.7                  & 31.0                  & 32.1                  \\ \hline

ADDS & \multirow{2}{*}{ov}   & \multirow{2}{*}{\textbf{36.56} }                 &  \multirow{2}{*}{\textbf{34.22}}                 &  \multirow{2}{*}{\textbf{36.65}}                 \\
(ViT-B-32, 224x224)   &                   &                   &                   \\ 

ADDS & \multirow{2}{*}{ov}  & \multirow{2}{*}{\textbf{39.01}}                 &  \multirow{2}{*}{\textbf{36.96}}                 &  \multirow{2}{*}{\textbf{39.28}} \\
(ViT-L-336, 336x336)   &                   &                   &                   \\ 
ADDS & \multirow{2}{*}{ov}  & \multirow{2}{*}{\textbf{42.67}}                 &  \multirow{2}{*}{\textbf{38.27}}                 &  \multirow{2}{*}{\textbf{40.49}} \\
(ViT-L-336, 448x448)   &                   &                   &                   \\ 


\hline                
\end{tabular}
}
\caption{Comparison of different methods on open-vocabulary multi-label classification for the NUS-WIDE dataset. In the \textit{Type} column, \textit{zsl} means the zero-shot learning setting, \textit{ov} means the open-vocabulary setting. Our ADDS method provides significantly better results than previous methods, including CONSE~\cite{norouzi2013zero}, LabelEM~\cite{akata2015label}, Fast0Tag~\cite{zhang2016fast}, One Attention per Label~\cite{kim2018bilinear}, One Attention per Cluster (M=10)~\cite{huynh2020shared}, LESA~\cite{huynh2020shared}, BiAM~\cite{narayan2021discriminative}, Generative ML-ZSL~\cite{gupta2021generative}, SDL~\cite{ben2021semantic}, and ML-Decoder~\cite{ridnik2021ml}.
}
\label{tab:nus}
\end{table}

Moreover, Table~\ref{tab:mscoco-zsl} shows the experimental results on the MSCOCO dataset. We make the data splitting in the following way: after sorting the class names in increasing alphabetical order, we select the first $65$ classes to be the seen classes, and the rest $15$ classes to be the unseen classes. The results show that our ADDS method can achieve much better results than the original ML-Decoder ($59.18\%$ vs. $30.69\%$), and when using the same backbone network, it still \textbf{significantly outperforms the ML-Decoder} ($54.52\%$ vs. $43.84\%$; note that ML-Decoder cannot take $448 \times 448$ images with ViT-L-336 as backbone).

\begin{table}[htb]
\centering
\small
\setlength{\tabcolsep}{1.4mm}{
\begin{tabular}{p{3.8cm}|c|c|c}

\hline
\multirow{2}{*}{Method}     & Input     & \multirow{2}{*}{mAP(\%)}                   & \multirow{2}{*}{F1(k=3)}                         \\
 &   Resolution  &                   &                          \\ \hline 

ML-Decoder (ViT-L-336, f) & {336x336}    & 43.84  &   35.08                                \\
ML-Decoder (ViT-L-336, uf) & {336x336}    & 43.75  &   17.09                                \\
ML-Decoder (TresNet-L)  &   {448x448}             &   30.69                 &  16.69                           \\
\hline

ADDS (ViT-L-336) &  336x336     & \textbf{54.52}     &    \textbf{51.52}                      \\
ADDS (ViT-L-336) &  448x448     &   \textbf{59.18}                 &  \textbf{77.34}                         \\
\hline                
\end{tabular}
}
\caption{Comparison on open-vocabulary multi-label classification for the MSCOCO dataset.
}
\label{tab:mscoco-zsl}
\end{table}

\definecolor{Gray}{gray}{0.9}

\subsection{Single-to-multi Label Classification}

\begin{table}[htbp]
\centering
\setlength{\tabcolsep}{1.4mm}{
\begin{tabular}{llccc}
\hline
\multicolumn{5}{c}{With overlapped classes}                                                                                                         \\ \hline
\multicolumn{1}{l|}{Model}      & \multicolumn{1}{l|}{Backbone}  & \multicolumn{1}{c|}{mAP(\%)} & \multicolumn{1}{c|}{F1(k=3)} & F1(k=5) \\ \hline
\multicolumn{1}{l|}{ML-Decoder} & \multicolumn{1}{l|}{TResNet-L} & \multicolumn{1}{c|}{41.60}   & \multicolumn{1}{c|}{17.80}   & 17.80   \\ \hline
\multicolumn{1}{l|}{ADDS}       & \multicolumn{1}{l|}{ViT-B-32}  & \multicolumn{1}{c|}{\textbf{59.27}}   & \multicolumn{1}{c|}{\textbf{45.88}}   & \textbf{45.88}   \\
\multicolumn{1}{l|}{ADDS}       & \multicolumn{1}{l|}{ViT-L-336} & \multicolumn{1}{c|}{\textbf{67.10}}   & \multicolumn{1}{c|}{\textbf{50.86}}   & \textbf{50.86}   \\ \hline
\multicolumn{5}{c}{Without overlapped classes}                                                                                                      \\ \hline
\multicolumn{1}{l|}{Model}      & \multicolumn{1}{l|}{Backbone}  & \multicolumn{1}{c|}{mAP(\%)} & \multicolumn{1}{c|}{F1(k=3)} & F1(k=5) \\ \hline
\multicolumn{1}{l|}{ML-Decoder} & \multicolumn{1}{l|}{TResNet-L} & \multicolumn{1}{c|}{38.37}        & \multicolumn{1}{c|}{6.90}        &   6.90      \\ \hline
\multicolumn{1}{l|}{ADDS}       & \multicolumn{1}{l|}{ViT-B-32}  & \multicolumn{1}{c|}{\textbf{64.32}}        & \multicolumn{1}{c|}{\textbf{30.90}}        & \textbf{30.90}        \\
\multicolumn{1}{l|}{ADDS}       & \multicolumn{1}{l|}{ViT-L-336} & \multicolumn{1}{c|}{\textbf{69.60}}        & \multicolumn{1}{c|}{\textbf{33.16}}        &  \textbf{33.16}       \\ \hline
\end{tabular}
}
\caption{Comparison on single-to-multi label classification. Models are trained on ImageNet-1k and tested on MS-COCO. Our method greatly outperforms ML-Decoder.}
\label{tab:MSCOCO-single-to-mul}
\end{table}

\begin{table}[htbp]
\centering
\setlength{\tabcolsep}{1.4mm}{
\begin{tabular}{llccc}
\hline
\multicolumn{5}{c}{With overlapped classes}                                                                                                         \\ \hline
\multicolumn{1}{l|}{Model}      & \multicolumn{1}{l|}{Backbone}  & \multicolumn{1}{c|}{mAP(\%)} & \multicolumn{1}{c|}{F1(k=3)} & F1(k=5) \\ \hline
\multicolumn{1}{l|}{ML-Decoder} & \multicolumn{1}{l|}{TResNet-L} & \multicolumn{1}{c|}{14.15}   & \multicolumn{1}{c|}{7.07}    & 7.30    \\ \hline
\multicolumn{1}{l|}{ADDS}       & \multicolumn{1}{l|}{ViT-B-32}  & \multicolumn{1}{c|}{\textbf{27.34}}   & \multicolumn{1}{c|}{\textbf{20.39}}   & \textbf{20.39}   \\
\multicolumn{1}{l|}{ADDS}       & \multicolumn{1}{l|}{ViT-L-336} & \multicolumn{1}{c|}{\textbf{31.07}}   & \multicolumn{1}{c|}{\textbf{24.69}}   & \textbf{24.69}   \\ \hline
\multicolumn{5}{c}{Without overlapped classes}                                                                                                      \\ \hline
\multicolumn{1}{l|}{Model}      & \multicolumn{1}{l|}{Backbone}  & \multicolumn{1}{c|}{mAP(\%)} & \multicolumn{1}{c|}{F1(k=3)} & F1(k=5) \\ \hline
\multicolumn{1}{l|}{ML-Decoder} & \multicolumn{1}{l|}{TResNet-L} & \multicolumn{1}{c|}{13.19}        & \multicolumn{1}{c|}{6.26}        &  6.27       \\ \hline
\multicolumn{1}{l|}{ADDS}       & \multicolumn{1}{l|}{ViT-B-32}  & \multicolumn{1}{c|}{\textbf{26.96}}        & \multicolumn{1}{c|}{\textbf{16.07}}        & \textbf{16.07}        \\
\multicolumn{1}{l|}{ADDS}       & \multicolumn{1}{l|}{ViT-L-336} & \multicolumn{1}{c|}{\textbf{30.66}}        & \multicolumn{1}{c|}{\textbf{19.18}}        & \textbf{19.18}        \\ \hline
\end{tabular}
}
\caption{Comparisons of single-to-multi label classification task which is trained on ImageNet-1k and tested on NUS-WIDE dataset. Our method shows the SOTA performance.}
\label{tab:NUS-single-to-mul}
\end{table}

We compare our approach with the previous method ML-Decoder in an extreme case of open-vocabulary multi-label classification, where models are trained on the single-label ImageNet-1k dataset and tested on the multi-label MS-COCO and NUS-WIDE datasets. We show two cases, depending on whether the testing dataset contains the overlapped classes with ImageNet-1k or not. The results are shown in Tables~\ref{tab:MSCOCO-single-to-mul} and~\ref{tab:NUS-single-to-mul}. Our approach greatly outperforms ML-Decoder, despite our model uses lower-resolution images ($224 \times 224$ for ViT-B-32 backbone and $336 \times 336$ for ViT-L-336 backbone) than ML-Decoder ($448 \times 448$). This again shows that \textbf{our approach greatly outperforms previous methods in single-to-multi label classification}.



\subsection{Additional Experiments}
\label{sec:additional_experiments}

\subsubsection{Conventional Multi-label Classification}

\begin{table}[]
\centering
\vspace{-0.2cm}
\begin{tabular}{l|l|c|c}
\hline
\multicolumn{1}{c|}{\multirow{2}{*}{Model}} & \multicolumn{1}{c|}{\multirow{2}{*}{Backbone}} & Input      & \multirow{2}{*}{mAP(\%)} \\
\multicolumn{1}{c|}{}                       & \multicolumn{1}{c|}{}                          & Resolution &                          \\ \hline
ML-GCN                                      & ResNet101                                      & 448x448    & 83.0                     \\
KSSNET                                      & ResNet101                                      & 448x448    & 83.7                     \\
SSGRL                                       & ResNet101                                      & 576x576    & 83.8                     \\
MS-CMA                                      & ResNet101                                      & 448x448    & 83.8                     \\
ASL                                         & TResNet-L                                      & 448x448    & 88.4                     \\
Q2L                                         & TResNet-L                                      & 448x448    & 89.2                     \\
ML-Decoder                                  & TResNet-M                                      & 224x224    & 84.2                     \\
ML-Decoder                                  & TResNet-L                                      & 448x448    & 90.1                     \\
ML-Decoder                                  & TResNet-XL                                     & 640x640    & 91.4                     \\
ML-Decoder                                  & ViT-L-336 (uf)                                    & 336x336    & 88.5                     \\
ML-Decoder                                  & ViT-L-336 (f)                                    & 336x336    & 90.6                     \\ \hline
ADDS                                        & ViT-L                                          & 224x224    & \textbf{89.82}                    \\
ADDS                                        & ViT-L-336                                      & 336x336    & \textbf{91.76}                    \\
ADDS                                        & ViT-L-336                                      & 640x640    & \textbf{93.41}                    \\
ADDS                                        & ViT-L-336                                      & 1344x1344  & \textbf{93.54}                        \\ \hline
\end{tabular}
\vspace{0.1cm}
\caption{Comparison on conventional multi-label classification for the MS-COCO dataset. Our ADDS approach shows significant improvement over previous methods, including ML-GCN~\cite{chen2019multi}, KSSNET~\cite{liu2018multi}, SSGRL~\cite{chen2019learning}, MS-CMA~\cite{you2020cross}, ASL~\cite{ben2020asymmetric}, Q2L~\cite{liu2021query2label}, and ML-Decoder~\cite{ridnik2021ml}.}
\label{tab:MSCOCO}
\end{table}

Apart from the open-vocabulary multi-label classification, we are also curious about how our model works on conventional multi-label classification. We conduct experiments on the MS-COCO dataset, and the results are shown in Table~\ref{tab:MSCOCO}. We run a few variations of our approach and compare them with a number of baselines. 
We first use the ViT-L backbone~\cite{dosovitskiy2020image} 
to test on $224 \times 224$ resolution images. We observe that our approach achieves an mAP of $89.82\%$, which is $5.6$ points higher than the previous SOTA method ML-Decoder~\cite{ridnik2021ml} on $224 \times 224$ images. We then switch to a larger image encoder ViT-L-336 for image resolution of $336 \times 336$. Our approach achieves an mAP of $91.76\%$, which is even better than what ML-Decoder can achieve on higher-resolution images of $640 \times 640$. 
And when we train our model with ViT-L-336 on $640 \times 640$ images, the mAP reaches $93.41\%$, $2.0$ points higher than ML-Decoder at the same resolution. With Pyramid-Forwarding, our model can also be deployed efficiently on an even higher resolution of $1344 \times 1344$ with the model ViT-L-336 pre-trained on $336 \times 336$ resolution, and our approach can achieve an mAP of $93.54\%$ (with much less computational cost than pre-training a model on $1344 \times 1344$ images). 

\subsubsection{Ablation Study: Effectiveness of DM-Decoder}
We validate the effectiveness of our Dual-Modal decoder (DM-Decoder) design. In particular, we conduct the open-vocabulary multi-label classification experiments on NUS-WIDE. We replace the DM-Decoder in our approach with the decoder layer in the previous SOTA ML-Decoder under various number of stacking layers. The image resolution is $336 \times 336$, and all image encoder is chosen as ViT-L-336. As shown in Table~\ref{tab:DM-decoder-zsl}, DM-Decoder significantly outperforms the decoder design in ML-Decoder in almost all cases, showing its effectiveness.

\begin{table}[htbp]
\centering
\small
\setlength{\tabcolsep}{1.4mm}{
\vspace{-0.3cm}
\begin{tabular}{l|c|c|c}
\hline
\multicolumn{1}{c|}{\multirow{2}{*}{Model}}  & \multirow{2}{*}{mAP(\%)} & \multirow{2}{*}{F1 (k=3)} & \multirow{2}{*}{F1 (k=5)}\\
\multicolumn{1}{c|}{}                       & \multicolumn{1}{c|}{} &\multicolumn{1}{c|}{}        &  \multicolumn{1}{c}{}      \\ \hline

ADDS+ML-Decoder$\times 1$                 & 36.15    & 32.45 & 35.50  \\
ADDS+ML-Decoder$\times 3$                 & 36.35    & 31.33 & 34.89  \\
ADDS+ML-Decoder$\times 6$                 & 36.34    & 29.83 & 33.56  \\
\hline
ADDS+DM-Decoder$\times 1$                 & \textbf{36.88}    & \textbf{32.95} & 35.48  \\
ADDS+DM-Decoder$\times 3$                 & \textbf{38.68}    & \textbf{34.46} & \textbf{37.50}  \\
ADDS+DM-Decoder$\times 6$                 & \textbf{39.01}    & \textbf{36.96} & \textbf{39.28}  \\
\hline
\end{tabular}
}
\vspace{0.1cm}
\caption{Comparison between DM-Decoder and ML-Decoder on NUS-WIDE for open-vocabulary multi-label classification.}
\vspace{-0.8cm}
\label{tab:DM-decoder-zsl}
\end{table}

\subsubsection{Ablation Study: Full Pyramid-Forwarding vs. Single-layer Pyramid-Forwarding}
We then evaluate the effectiveness of Pyramid-Forwarding by comparing the results of using only a single layer of Pyramid-Forwarding versus using full Pyramid-Forwarding on MS-COCO. We choose this dataset since it has higher resolution and more suitable for comparing on different resolutions. In Table~\ref{tab:abla-infer-log}, the second column shows the level index in Pyramid-Forwarding. The first line shows the result of using the model on $336 \times 336$ images without Pyramid-Forwarding. The second line shows the model with Pyramid-Forwarding on $1344 \times 1344$ images, but only with the level of highest resolution (third level) and cutting into $16$ patches. The third line shows the model with full Pyramid-Forwarding on $1344 \times 1344$ resolution images. We can see that using full Pyramid-Forwarding cannot provides much more performance boost.

\begin{table}[htbp]
\centering
\small
\vspace{-0.2cm}
\begin{tabular}{l|c|c|c}
\hline
\multirow{2}{*}{Model} & \multirow{2}{*}{Level} & Image  & \multirow{2}{*}{mAP(\%)}\\
          &                & Resolution &                          \\ \hline

ADDS                &      [0]                  & 336x336                                     & 91.76        \\ 
ADDS                &       [2]                 & 1344x1344                                      & 91.79       \\
ADDS                &       [0,1,2]                 & 1344x1344                                     & 93.54       \\
\hline
\end{tabular}
\vspace{0.1cm}
\caption{Full vs. single-layer Pyramid-Forwarding.}
\vspace{-0.6cm}
\label{tab:abla-infer-log}
\end{table}

\subsubsection{Ablation Study: Impact of the Number of Training Classes}
We claim that increasing the number of available training classes can benefit the single-to-multi label classification, and we validate this through training our model (on ViT-L-336) with the selective language supervision technique on the ImageNet-21k dataset. For a fair comparison, we filter out the overlapped classes with NUS-WIDE in ImageNet-1k, and we select the first 15k classes in ImageNet-21k without the overlapped classes with NUS-WIDE. We select 100 images for each class, so that the selected dataset contains 1.3M images which is the same level as ImageNet-1k (1.3M). The result of training on ImageNet-1k is shown on the first line of Table~\ref{tab:ImageNet+NUS}, and the second line shows the result of ImageNet-21k. With the number of available training classes becomes 15 times than before, the model mAP increases 6.9 points.

\begin{table}[htbp]
\centering
\setlength{\tabcolsep}{1.4mm}{
\begin{tabular}{l|l|c|c|c}
\hline
Train Dataset & Backbone & mAP(\%) & F1(k=3) & F1(k=5)\\
\hline

ImageNet-1k  & ViT-L-336              & 31.02     &     24.98   & 24.98    \\
ImageNet-21k & ViT-L-336              & \textbf{37.92}     &     \textbf{39.82}   & \textbf{40.39}    \\
\hline
\end{tabular}
}
\caption{Comparisons of our method trained on the ImageNet-1k vs. ImageNet-21k dataset (filter the overlapped classes with NUS-WIDE) and tested on the NUS-WIDE dataset, with 336x336 resolution and ViT-L-336 backbone.}
\vspace{-0.6cm}
\label{tab:ImageNet+NUS}
\end{table}

\subsubsection{Experiments with Other VLP Models}
Finally, we are also curious about how other VLP models perform under our method. We consider two examples BLIP~\cite{li2022blip} and SLIP~\cite{mu2021slip}. They both have the contrastive loss similar to CLIP to ensure an alignment between the visual and textual features. In the second line of Table~\ref{tab:otherVLP}, the BLIP model with its ViT-L image encoder on $224 \times 224$ images shows a good result of $35.15\%$. SLIP in the third line shows similar performance. However, when we use a BLIP model that is fine-tuned on MS-COCO without the contrastive loss to ensure the alignment, the mAP quickly goes down to $2.52\%$. This strongly shows that the correlation between visual and textual embedding plays an important role in providing the performance boost, instead of the image or text encoder itself.

\begin{table}[htbp]
\small
\centering
\begin{tabular}{c|l|c|c}
\hline
Image/Text & \multirow{2}{*}{Backbone} & Image  & \multirow{2}{*}{mAP(\%)}\\
 Encoder                       &                           & Resolution &                          \\ \hline

BLIP   & ViT-L(COCO)     &     384x384   & 2.52    \\
BLIP   & ViT-L     &     224x224   & 35.15    \\
SLIP   & ViT-L     &     224x224   & 34.15    \\
\hline
\end{tabular}
\vspace{0.1cm}
\caption{Applying other VLP models for alignment to our method on NUS-WIDE open-vocabulary multi-label classification task.}
\vspace{-0.4cm}
\label{tab:otherVLP}
\end{table}

%% file: sections/conclusion.tex
\section{Conclusion}
In this paper, we present a novel open-vocabulary multi-label classification framework called Aligned Dual moDality ClaSsifier (ADDS). The framework is based on textual-visual alignment and leverages a novel Dual-Modal Decoder design and a Pyramid-Forwarding technique. 
Our method significantly outperforms all previous results and becomes the state-of-the-art method on open-vocabulary multi-label classification, single-to-multi label classification, and also conventional multi-label classification tasks. 
